\documentclass{article}
\usepackage{ijcai25}

\usepackage{times}
\usepackage{soul}
\usepackage{url}
\usepackage[hidelinks]{hyperref}
\usepackage[utf8]{inputenc}
\usepackage[small]{caption}
\usepackage{graphicx}
\usepackage{amsmath}
\usepackage{amsthm}
\usepackage{booktabs}
\usepackage{algorithm}
\usepackage{algorithmic}
\usepackage[switch]{lineno}
\usepackage{multirow}
\usepackage{xurl}
\usepackage{CJKutf8}
\usepackage{bm}

\usepackage{amssymb}

\title{SMILE: a Scale-aware Multiple Instance Learning Method for Multicenter STAS  Lung Cancer Histopathology Diagnosis}

\author{
Liangrui Pan$^1$
\and
Xiaoyu Li$^1$\and
Yutao Dou$^1$\And
Qiya Song$^2$\and
Jiadi Luo$^{3}$\And
Qingchun Liang$^3$\And
Shaoliang Peng$^1$\\
\affiliations
$^1$College of Computer Science and Electronic Engineering, Hunan University, Changsha 410082, China\\
$^2$College of Information Science and Engineering, Hunan Normal University, Changsha 410082, China\\
$^3$Department of Pathology, The Second Xiangya Hospital, Central South University, Changsha, 410011, Hunan, China\\
\emails
\{panlr,hnulixy, ytdou, sqyunb\}@hnu.edu.cn,
\{503079, jiadiluo\}@csu.edu.cn, 
}

\begin{document}
\begin{CJK}{UTF8}{gbsn}
\maketitle

\begin{figure*}[htbp]
	\centering
	\includegraphics[width=1\textwidth]{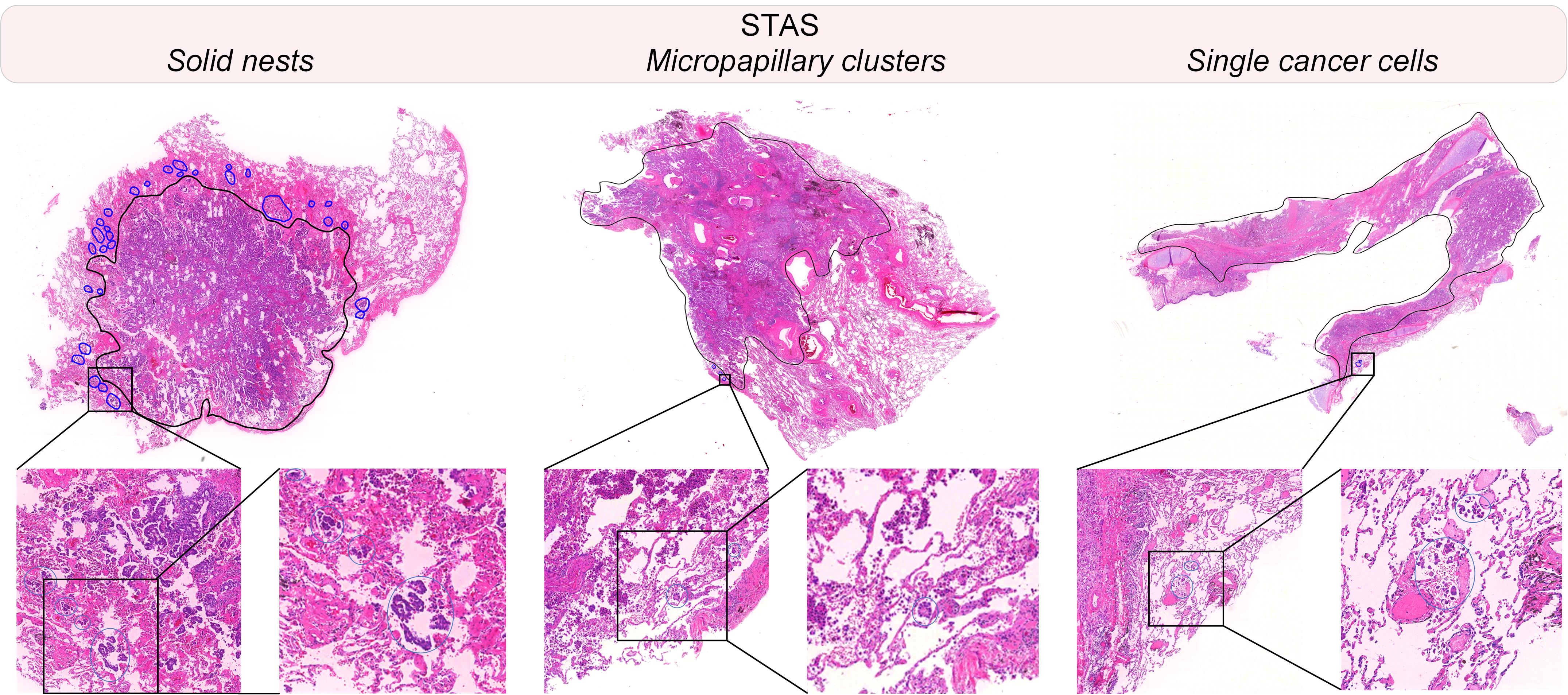}
	\caption{Three common pathological features of STAS in lung cancer histopathology images. STAS is mainly distributed outside the main tumor body in the form of solid cell nests, micropapillary clusters, and single cancer cells.}
	\label{fig:stas}
\end{figure*}

\begin{abstract}
    Spread through air spaces (STAS) represents a newly identified aggressive pattern in lung cancer, which is known to be associated with adverse prognostic factors and complex pathological features. Pathologists currently rely on time-consuming manual assessments, which are highly subjective and prone to variation. This highlights the urgent need for automated and precise diagnostic solutions. 2,970 lung cancer tissue slides are comprised from multiple centers, re-diagnosed them, and constructed and publicly released three lung cancer STAS datasets: STAS\_CSU (hospital), STAS\_TCGA, and STAS\_CPTAC. All STAS datasets provide corresponding pathological feature diagnoses and related clinical data. To address the bias, sparse and heterogeneous nature of STAS, we propose an scale-aware multiple instance learning(SMILE) method for STAS diagnosis of lung cancer. By introducing a scale-adaptive attention mechanism, the SMILE can adaptively adjust high-attention instances, reducing over-reliance on local regions and promoting consistent detection of STAS lesions. Extensive experiments show that SMILE achieved competitive diagnostic results on STAS\_CSU, diagnosing 251 and 319 STAS samples in CPTAC and TCGA, respectively, surpassing clinical average AUC. The 11 open baseline results are the first to be established for STAS research, laying the foundation for the future expansion, interpretability, and clinical integration of computational pathology technologies. The datasets and code are available at \url{https://anonymous.4open.science/r/IJCAI25-1DA1}. 
   
\end{abstract}

\section{Introduction}

Spread Through Air Spaces (STAS) is recognized as a newly described invasive pattern in lung cancer. Onozato \textit{et al.} \cite{onozato2013tumor} first noted tumor cells occupying alveolar spaces in 2013. The World Health Organization (WHO) later defined STAS as tumor cells (e.g., micropapillary clusters, solid nests, or single cells) that spread within air spaces beyond the main tumor \cite{chae2021prognostic}. Using three-dimensional reconstruction, tumor islands initially appearing separate were found to be connected to the main tumor on different planes \cite{warth2015prognostic}, implying an aggressive invasion mode correlated with higher tumor grade, KRAS mutations, and poorer recurrence-free survival. Numerous studies have shown that STAS frequently coexists with other high-risk pathological features, including pleural invasion, vascular invasion, larger tumor size, and higher stage, which collectively indicate a worse prognosis \cite{eguchi2019lobectomy,shiono2016spread}.

Histopathological images remain the gold standard for diagnosing STAS. Nevertheless, identification of STAS relies heavily on pathologists' manual observation, leading to considerable subjectivity and dependence on individual expertise. The resultant workload can be overwhelming when processing large-scale slides, which risks both extended turnaround times and potential diagnostic inconsistencies. STAS status is also pivotal for surgical decision-making, such as choosing between lobectomy and sub-lobectomy, as intraoperative removal of STAS-affected regions improves patient survival \cite{khalil2023analysis,zhou2022assessment}. However, manual diagnostic approaches reportedly achieve about 74\% accuracy (AUC = 0.67) \cite{villalba2021accuracy,wang2023deep}, underscoring the urgent need for more objective and efficient diagnostic methods.

Artificial intelligence has shown tremendous promise in assisting pathological diagnoses, with notable successes in tumor classification, lesion boundary detection, and survival prediction. However, The diagnosis of STAS remains challenging. As an invasive pattern that has only recently been incorporated into pathology reports, STAS lacks a universally accepted diagnostic standard across clinical and pathology communities. This inconsistency leads to variations in diagnostic agreement and sensitivity among different institutions. Additionally, the availability of high-quality, multicenter annotated pathology datasets for STAS have not yet appeared, restricting the breadth and depth of related research.  As illustrated in Figure~\ref{fig:stas}, STAS is characterized by tumor cell dissemination along the small airways, often presenting as scattered, minute lesions. These lesions vary in size, morphology, and exhibit significant heterogeneity, making comprehensive identification through conventional manual examination particularly difficult. Traditional pathological diagnostic approaches have inherent limitations in detecting STAS, especially in distinguishing subtle lesions and irregular morphological patterns, which can introduce diagnostic errors. Furthermore, accurate STAS assessment requires an in-depth understanding of tumor dissemination pathways within air spaces, further complicating the diagnostic process for pathologists. These challenges highlight the urgent need to refine diagnostic strategies to enhance the consistency, efficiency, and interpretability of STAS identification.

To address the aforementioned challenges, we downloaded all lung cancer histopathological images from The Cancer Genome Atlas (TCGA) and the Clinical Proteomic Tumor Analysis Consortium (CPTAC) projects. Three pathologists cross-diagnosed the STAS status and the types of dissemination foci for each slide, establishing the STAS\_TCGA and STAS\_CPTAC datasets. Importantly, we collected and curated histopathological images from STAS lung cancer patients at the Second  Xiangya Hospital of Central South University and constructed the STAS\_CSU dataset. Given the sparsity and heterogeneity of STAS pathological features, which typically appear as isolated cells or small clusters beyond the primary tumor body and are often difficult to detect in histopathological images, we propose SMILE, a novel scale-aware multiple instance learning method designed for STAS whole-slide image (WSI) classification. This approach introduces a scale-aware strategy to reduce the model's over-reliance on high-attention instances, thereby enhancing its ability to capture sparse and heterogeneous features and improving STAS recognition accuracy. Our main contributions are as follows:
\begin{itemize}
	\item This study conducted STAS diagnosis on lung cancer patients from TCGA and CPTAC, and for the first time, constructed and publicly released three STAS datasets: STAS\_CSU, STAS\_TCGA, and STAS\_CPTAC, comprising a total of 2,970 histopathological images.
	\item We introduce a scale-adaptive attention mechanism that optimizes the SMILE to focus more evenly on instances in the bag by dynamically adjusting the attention to high-attention instances, thus improving the accuracy of STAS prediction in histopathology images.
	\item We conduct benchmark evaluations of 11 multi-instance learning algorithms across Three STAS datasets, delivering comprehensive baseline results to advance research on STAS-assisted diagnosis.  
\end{itemize}


\section{Related Work}

\subsection{Lung Cancer Histopathology Image Datasets}
Currently, publicly accessible lung cancer histopathological image data mainly come from large-scale projects such as TCGA, CPTAC, and Cancer Digital Slide Archive (CDSA), as well as various datasets on the Kaggle platform \cite{gutman2017digital,borkowski2019lung,heath2021nci,lu2017spread}. The TCGA initiative provides WSIs and corresponding molecular data for both lung adenocarcinoma (LUAD) and squamous cell carcinoma (LUSC) via the Genomic Data Commons portal \cite{han2021tumor}. The CPTAC program offers digital pathology images for multiple cancer types, including lung cancer, while integrating proteomic data to facilitate multimodal research. On Kaggle, a well-known dataset titled “Lung and Colon Cancer Histopathological Images” includes hundreds of samples, 750 of which (250 benign lung tissue, 250 LUAD, and 250 LUSC) are labeled and suitable for classification or segmentation\footnote{\url{https://www.kaggle.com/datasets/andrewmvd/lung-and-colon-cancer-histopathological-images}}. The CDSA also provides digital pathology slides; notably, the National Lung Screening Trial subset features 1,225 high-quality WSIs\footnote{\url{https://cdas.cancer.gov/datasets/nlst/}}. In addition, certain national cancer centers possess lung cancer histopathological datasets for domestic research, typically containing image, radiological, molecular, and clinical information. Although STAS has been repeatedly noted by the WHO as an invasive malignancy manifestation, most publicly available datasets do not specify STAS annotations in their diagnostic records. To address this gap, we re-invited three experienced pathologists to cross-diagnosis STAS labels for all lung cancer WSIs in the TCGA and CPTAC datasets, thus enriching the data foundation for STAS research and intelligent diagnostic modeling.

\subsection{STAS Diagnosis}
STAS in the lung correlates strongly with the surgical approach and poor prognosis in early-stage lung adenocarcinoma \cite{lin2024ct,feng2024deep}. Preoperative STAS prediction is therefore critical for surgical planning, yet STAS detection remains challenging due to false positives, low inter-observer agreement, and limited quantitative analysis. One STAS-DL model extracted solid-component features through a solid component gating (SCG) mechanism and achieved an AUC of 0.82 and 74\% accuracy, surpassing both STAS-DL without SCG (70\% accuracy) and physician performance (AUC = 0.68) \cite{lin2024ct}. Another ResNet-18-based deep learning method yielded an AUC of 0.841. A hybrid model combining deep learning and radiomics improved performance by 3.50\% and 4.60\% compared to either component alone. The STASNet approach computes semi-quantitative STAS parameters (density and distance) and obtained 0.93 patch-level detection accuracy with 0.72 AUC at the WSI \cite{feng2024deep}. Similarly, researchers have leveraged machine learning and deep learning to build STAS prediction models using radiological-histological features (AUC = 0.764 in training, 0.776 in testing), while integrating clinical data raised the AUC to 0.878. A ResNet50-based model further reached 0.918 AUC. In terms of graph modeling, Cen et al. introduced Ollivier-Ricci curvature-based graph theory to enhance accuracy and explainability via primary tumor margin features \cite{cen2024orcgt}. Pan et al. then proposed VERN, a feature-interactive siamese encoder that performs effectively on both frozen sections (FSs) and paraffin sections (PSs) \cite{pan2024feature}. Although most deep learning-based STAS diagnostics currently focus on radiomics and well-designed STAS datasets,, their performance has yet to reach clinical-grade levels. In contrast, computational pathology for STAS remains a burgeoning focus of research.

\subsection{Multiple Instance Learning}
Given the importance of weakly supervised learning in digital pathology, Multiple instance learning (MIL) is widely employed for WSI analysis using only slide-level labels without labor-intensive pixel annotations. MIL models typically fall into two groups: one predicts directly at the instance level and aggregates these outputs for bag-level decisions \cite{li2021dual,wang2019weakly,hou2016patch,shao2021transmil}; the other extracts instance features, then combines these into a bag representation for classification \cite{yao2020whole,zhang2022dtfd,zhao2020predicting,ilse2018attention,lin2023interventional}. While mean-pooling or max-pooling are straightforward ways to aggregate instance probabilities, they often perform worse than bag embedding methods \cite{zhang2022dtfd,wang2022transformer}. Bag embedding learns a high-level representation for the entire bag, producing more robust features. Most bag-embedding-based techniques employ attention mechanisms, as seen in ABMIL, which sums instance features with learned attention weights. These weights can be determined by a side network \cite{ilse2018attention}, by cosine distance to key instances \cite{hou2016patch}, or by a transformer architecture that encodes inter-instance relationships \cite{shao2021transmil}. Various extensions of ABMIL and TransMIL have pushed MIL research further in pathology image analysis \cite{shi2024cod,lu2023visual,jaume2024multistain,wang2022transformer}. Given that STAS histopathological images typically exhibit high resolution, sparsity, complex textures, and pronounced heterogeneity, we proposes a MIL-based diagnostic approach for STAS using a scale-adaptive attention mechanism to improve WSI analysis and advance the clinical application of computational pathology.

\section{Lung Cancer STAS Datasets}

\subsection{Dataset Construction}

\begin{figure}[htbp]
	\centering
	\includegraphics[width=0.48\textwidth]{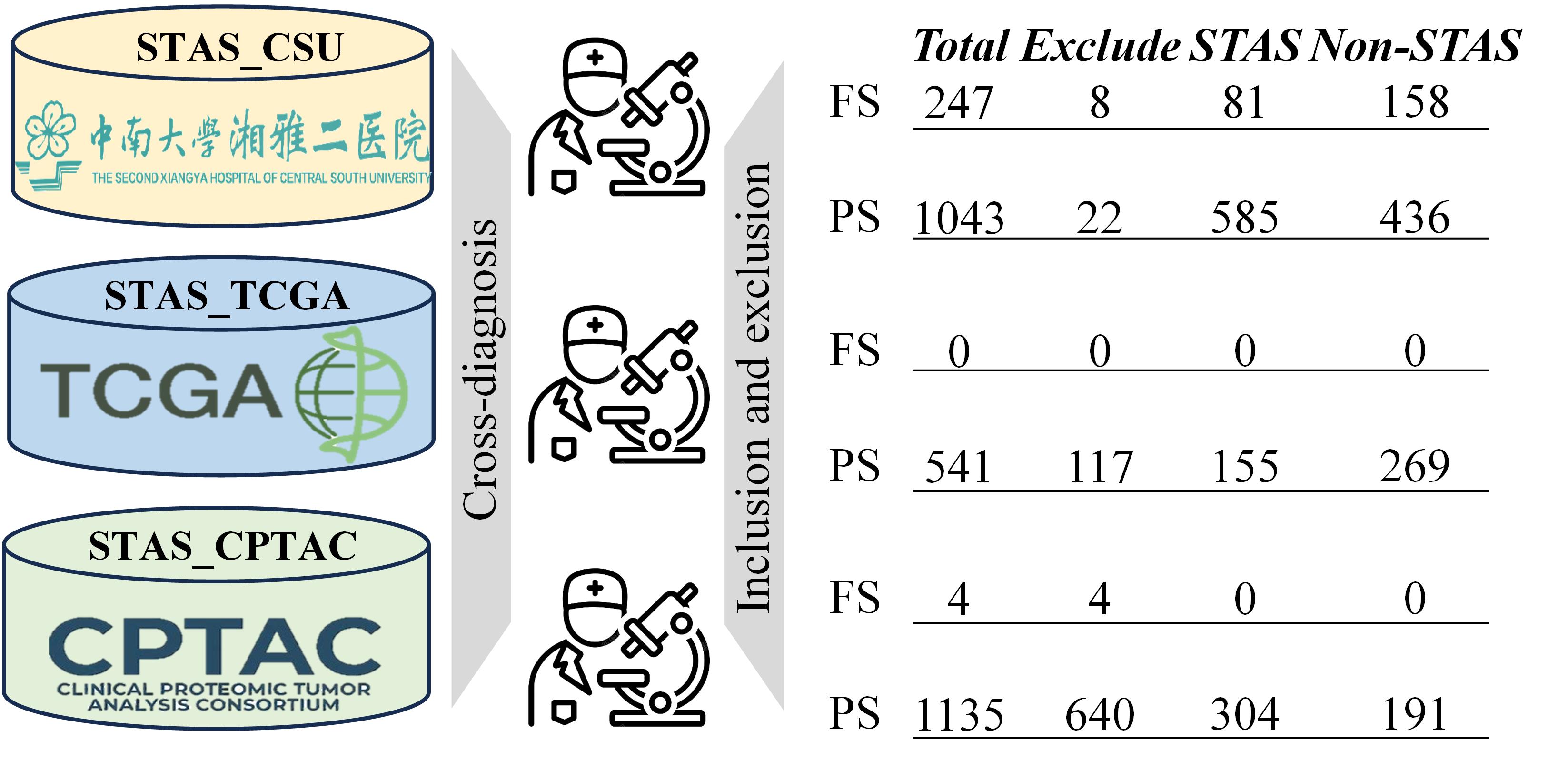}
	\caption{The process of constructing the three STAS datasets.}
	\label{fig:data}
\end{figure}

As illustrated in Figure~\ref{fig:data}, We compiled three STAS datasets for this study, comprising STAS\_CSU from the Second Xiangya Hospital, STAS\_TCGA from the U.S. National Cancer Institute, and STAS\_CPTAC from the CPTAC. All histopathological data were reviewed by three experienced pathologists, who determined the STAS label for each WSI by observing the tissue under a microscope. Following a double-blind experimental protocol, we employed cross-diagnosis to obtain accurate labels and minimize pathologist bias, as well as reduce the risk of missed or incorrect diagnoses. Based on our research objectives, we only included WSIs from patients meeting the following criteria: (1) diagnosed with LUAD, (2) availability of corresponding routine pathological slides containing both the primary tumor and adjacent non-tumor tissue, (3) detailed TNM staging, (4) high-quality slides without bending, wrinkling, blurring, or color distortion, and (5) exclusion of slides containing the primary tumor but lacking adjacent non-tumor tissue. Below, we provide a detailed description of our dataset.

\noindent\textbf{STAS\_CSU:} From April 2020 to December 2023, we selected 356 patients at the Second Xiangya Hospital who underwent pulmonary nodule resection and were diagnosed with lung cancer (particularly those with STAS) to form a retrospective lung cancer cohort. We comprehensively collected each patient's clinical and pathological data, including age, tumor size, lymph node metastasis, distant metastasis, clinical stage, recurrence time and status, as well as survival time and status. Two experienced pathologists independently reviewed the pathology data for every patient, including frozen and paraffin-embedded hematoxylin and eosin (H\&E) stained slides, immunohistochemical (IHC) slides, confirming the presence or absence of STAS, the specific pathological subtype of any disseminated foci, the detailed histological subtype of lung cancer, and the expression of key proteins (PD-L1, TP53, Ki-67, and ALK). Within this cohort, there were 150 non-STAS patients and 206 STAS patients. Each patient's tumor specimen was sectioned by pathologists into multiple paraffin blocks, each corresponding to multiple H\&E slides. In total, we collected and scanned 1{,}290 FSs and PSs and 1{,}436 IHC slides. Of these, 247 FSs comprised 81 STAS and 158 non-STAS FSs, while 1{,}043 PSs contained 585 STAS and 436 non-STAS PSs. All FSs and PSs were digitized into WSI. 

\noindent \textbf{STAS\_TCGA:} We downloaded relevant LUAD WSIs from the TCGA website\footnote{\url{https://portal.gdc.cancer.gov/}}. Based on our inclusion and exclusion standards, we collected 541 WSIs of PS from an unspecified number of patients. All WSIs underwent cross-diagnosis by three experienced pathologists to determine STAS status, type of dissemination foci, and tumor type. We found that STAS was positively correlated with the status and duration of survival. Finally, following the inclusion and exclusion criteria, the STAS\_TCGA dataset consists of 155 STAS WSIs and 269 non-STAS WSIs, along with corresponding patient survival times and statuses, and 117 WSIs were excluded.

\noindent \textbf{STAS\_CPTAC:} We obtained 1{,}139 WSIs from the CPTAC\footnote{\url{https://www.cancerimagingarchive.net/collection/cptac-luad/}}. According to our inclusion and exclusion criteria, 304 WSIs of PS were labeled as STAS, 191 were labeled as non-STAS, and 640 WSIs were excluded. The included image data was Cross-diagnosed by three pathologists to obtain the pathological type, including the STAS status. These images were then used to evaluate the performance of our model. Moreover, this dataset also provides multi-omics data and clinical data related to lung cancer.

\subsection{Data Preprocessing}
This stage primarily involves WSI digitization, followed by applying the OTSU algorithm to segment tissue regions, detect background, and identify blurred areas \cite{jothi2016effective}. All WSIs are automatically processed to generate thumbnails, masks, and overview images. Next, we segment each WSI (at 20$\times$ magnification) into patches of size 256$\times$256 pixels, recording the coordinates and position of each patch. To mitigate the impact of dataset quality on model generalization, we employ a GAN based on pathological image features to enhance the patches \cite{xue2021selective}.  

\section{Method}

\begin{figure*}[h!]
	\centering
	\includegraphics[width=0.95\textwidth]{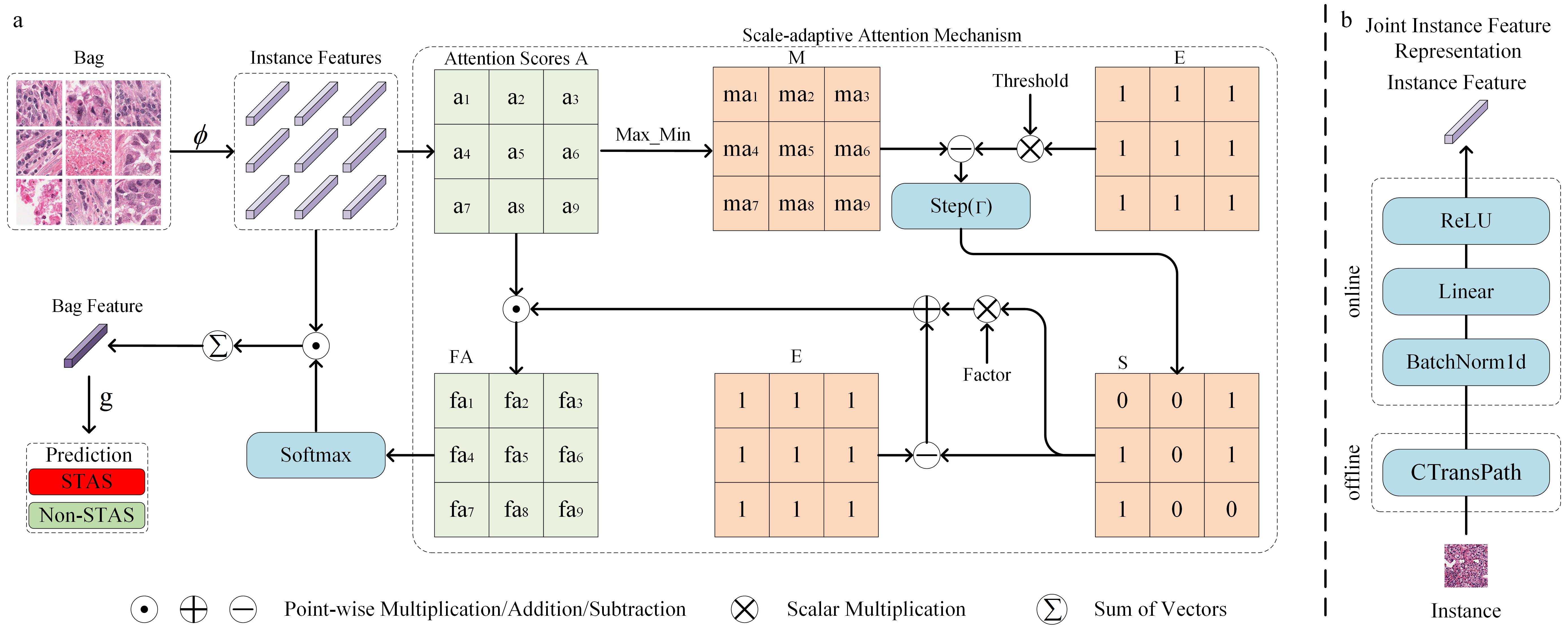}
	\caption{(a) Overall workflow of the proposed SMILE approach. (b) The process of feature preprocessing. We process the given bag through a joint feature representation module to transform them into instance features. These features are then processed through a scale-adaptive attention module to obtain scaled bag-level feature representations. Finally, the final STAS prediction results are obtained through the classifier $g$.}
	\label{fig:workflow}
\end{figure*}	
\subsection{Problem Definition}
In the MIL framework, each bag is treated as a labeled unit, while instances within the bag may possess different feature representations. Consider a binary classification task where a bag \(X = \{ x_1, x_2, \ldots, x_n \}\) contains \(n\) instances, and each instance \(x_i\) can be predicted as positive \((y_i = 1)\) or negative \((y_i = 0)\). If \emph{at least one} instance \(x_i\) in the bag is classified as positive \(\bigl(y_i = 1\bigr)\), the bag is labeled positive \(\bigl(Y = 1\bigr)\); otherwise, if all instances are negative \(\bigl(\sum y_i = 0\bigr)\), the bag is labeled negative \(\bigl(Y = 0\bigr)\). Formally,
\[
Y = 
\begin{cases}
	0 & \text{if} \ \sum_i y_i = 0, \\
	1 & \text{otherwise}.
\end{cases}
\tag{1}
\]

When applying the MIL framework to STAS prediction, each WSI is viewed as a bag labeled either STAS or non-STAS. Due to the high resolution and large size of a WSI, it cannot be directly fed into a MIL model. Thus, the WSI is usually divided into multiple smaller patches that serve as instances, each containing local information about the WSI. Within the MIL paradigm, if at least one instance \(x_i\) is identified as STAS (\(y_i = 1\)), the entire WSI (bag) is labeled STAS. If no instance is STAS (\(\forall \, i, \ y_i = 0\)), then the WSI is labeled non-STAS.

\subsection{Scale-aware Multiple Instance Learning}

As illustrated in Figure~\ref{fig:workflow}(a), our  proposed STAS prediction process based on scale-aware multiple instance learning (SMILE) involves three key steps:  
\textbf{(1)} \emph{Instance-Level Feature Extraction}: For each instance \(x_i\), extract a feature vector \(\varphi(x_i)\).  
\textbf{(2)} \emph{Feature Fusion}: Aggregate all instance-level features from bag \(X = \{ x_1, x_2, \dots, x_n \}\) into a single bag-level feature representation \(\phi(X)\).  
\textbf{(3)} \emph{Bag-Level Prediction}: Use a bag-level classifier \(g\) to predict the label \(\hat{Y}\) based on the fused feature \(\phi(X)\). Formally,
\begin{align}
	\hat{Y} 
	&= g\Bigl(\phi(\varphi(x_1),\, \varphi(x_2), \ldots,\varphi(x_n))\Bigr), 
	\nonumber \\
	&\quad \forall \, i \in \{1,2,\ldots,n\}.
	\tag{2}
\end{align}
\subsection{Instance-Level Feature Extraction}
Given a bag \(X = \{ x_1, x_2, \ldots, x_n \}\in \mathbb{R}^{n \times 256 \times 256 \times 3}\), we first extract a high-dimensional feature representation for each instance, capturing local information and properties. Formally,
\[
H = \varphi(X) = \{h_1, h_2, \ldots, h_n\},
\tag{3}
\]
where \(H \in \mathbb{R}^{n \times d}\) is the feature matrix of bag \(X\), and \(h_i\) is the feature vector of the \(i\)-th instance.

As illustrated in Figure~\ref{fig:workflow}(b), the feature extraction is divided into two stages: \emph{offline} and \emph{online}. During the \emph{offline} phase, we use a backbone network called \texttt{CTransPath}, which combines convolutional neural network and Transformer architectures \cite{wang2022transformer}. After pretraining \texttt{CTransPath} with semantic contrastive learning, its weights are frozen for all subsequent stages. The patches are fed into \texttt{CTransPath}, which subsequently outputs the feature vector representations of these patches. Formally,
\[
T = \texttt{CTransPath}(X).
\tag{4}
\]
In the \emph{online} phase, we apply three layers: \(\texttt{BatchNorm1d}\), \(\texttt{Linear}\), and \(\texttt{ReLU}\). These layers continue to be trainable, allowing fine-tuning of the extracted features:
\[
R = \texttt{BatchNorm1d}(T),
\tag{5}
\]
\[
H = \texttt{ReLU}\bigl(\texttt{Linear}(R)\bigr),
\tag{6}
\]
where \(T = \{t_1, t_2, \ldots, t_n\} \in \mathbb{R}^{n \times l}\) and \(R = \{r_1, r_2, \ldots, r_n\} \in \mathbb{R}^{n \times l}\).

\subsection{Scale-adaptive Instance Space}
During instance feature fusion, we introduce an attention module that takes the instance-level feature matrix \(H\) as input and outputs an attention weight for each instance:
\[
A = W_a \Bigl(\tanh\bigl(V_a H\bigr) \,\odot\, \sigma\bigl(U_a H\bigr)\Bigr) = \{a_1,a_2,\ldots,a_n\},
\tag{7}
\]
where \(A \in \mathbb{R}^{n \times 1}\) represents the unnormalized attention scores; \(V_a \in \mathbb{R}^{e \times d}\), \(U_a \in \mathbb{R}^{e \times d}\), and \(W_a \in \mathbb{R}^{1 \times e}\) are learnable matrices; \(\tanh(\cdot)\) and \(\sigma(\cdot)\) denote the tangent and sigmoid activation functions, respectively; \(\odot\) is the element-wise multiplication operator.


To enable the model to learn more generalizable feature representations, we designed a \emph{scale-adaptive attention mechanism} that scales high attention scores exceeding a certain threshold based on the attention scores \(A\). Specifically, after applying Max\_Min normalization to \(A\), we clamp the values exceeding a preset threshold by a factor, mitigating the model's over-dependence on certain high-attention instances. Mathematically,
\[
S = \Gamma\Bigl(\text{Max\_Min}(A) \,\ominus\, \bigl(\text{Threshold} \,\otimes\, E\bigr)\Bigr),
\tag{8}
\]
\[
\Gamma(x) =
\begin{cases}
	0 & \text{if } x<0,\\
	1 & \text{if } x \geq 0,
\end{cases}
\tag{9}
\]

\[
\text{Max\_Min}(x) = \frac{x - \min(x)}{\max(x) - \min(x)}
\tag{10}
\]

\begin{align}
\text{SA} 
	&= \text{Softmax}\Bigl(A \,\odot\, \bigl((E \ominus S)\,\oplus\,(\text{Factor}\,\otimes\,S)\bigr)\Bigr)
	\nonumber \\
	&= \{\,\text{sa}_1,\,\text{sa}_2,\,\ldots,\,\text{sa}_n\},
	\tag{11}
\end{align}
where \(E \in \mathbb{R}^{n \times 1}\) is a vector of ones, \(\otimes\) denotes element-wise multiplication by a scalar, and \(\oplus\), \(\ominus\) denote element-wise addition and subtraction, respectively. Threshold and Factor represent the threshold value and scaling factor, respectively.

Finally, we compute a global representation by aggregating all instance features with the scaled attention weights:
\[
z = \sum_{i=1}^n \text{sa}_i \,h_i.
\tag{12}
\]
For bag-level classification, we use a linear layer with a sigmoid activation:
\[
\hat{Y} = \sigma\bigl(\text{Linear}(z)\bigr).
\tag{13}
\]

\subsection{Loss Function}

To optimize the parameters of SMILE, we adopt the standard cross-entropy loss. Given a predicted probability distribution $\hat{y}$ and the ground-truth label $y \in \{0,1,\dots,C{-}1\}$ for $C$ classes, the cross-entropy loss is defined as
\begin{equation}
	\mathcal{L}_{\mathrm{CE}} 
	= - \sum_{c=0}^{C-1} y_{c} \log\big(\hat{y}_{c}\big),
	\tag{14}
\end{equation}
where $\hat{y}_{c}$ denotes the predicted probability for class $c$. For binary classification ($C=2$), this reduces to
\begin{equation}
	\mathcal{L}_{\mathrm{CE}} 
	= - \Big[y \log\big(\hat{y}\big)
	+ (1-y)\,\log\big(1 - \hat{y}\big)\Big].
	\tag{15}
\end{equation}
Minimizing $\mathcal{L}_{\mathrm{CE}}$ encourages the model to assign higher confidence scores to correct classes, thus improving overall classification performance.

\section{Experiments}

\subsection{Experimental Setups}
\noindent \textbf{Dataset setup.} We used three STAS datasets in total—STAS\_CSU, STAS\_TCGA, and STAS\_CPTAC—for training and evaluation. For each dataset, we applied five-fold cross-validation, partitioning the dataset into five subsets, training on four subsets and using the remaining subset for validation. After performing five-fold cross-validation, we obtained five best-performing prediction models. The experimental results use the average value of five-fold cross-validation as the statistical value.

\noindent \textbf{Evaluation metrics.} We utilize the area under the receiver operating characteristic curve (AUC), accuracy, precision, recall, and F1-Score as evaluation metrics to comprehensively assess model performance from multiple perspectives. 

\noindent \textbf{Implementation details.} All models were trained on the PyTorch framework using two NVIDIA RTX 4090 GPUs. The experiments employed the Ranger optimizer to adjust model parameters, with a learning rate of $2e^{-4}$, weight decay of $1e^{-5}$, over 100 training epochs, and a batch size of 12. The threshold was set to 0.5 with a scaling factor of 0.5. $l$ was 768, the feature compression dimension $d$ was 256, and the attention mechanism parameter $e$ was 64.

\subsection{Comparison of STAS Diagnosis Methods}
In our experiment, we considered 11 STAS diagnostic methods. Since each method needs to be run on a multicenter STAS dataset, we only selected those methods that have publicly available code and corresponding MIL models.


\begin{itemize}
	\item \textbf{Maxpooling} represents a slide by selecting the instance with the maximum activation, thereby mimicking the focus on the most prominent lesion.
	\item \textbf{Meanpooling} aggregates all instance features by computing their mean, thus treating each patch equally in the overall representation.
	\item \textbf{ABMIL} \cite{ilse2018attention} employs an attention mechanism to assign weights to instances, effectively prioritizing diagnostically relevant regions.
	\item \textbf{TransMIL} \cite{shao2021transmil} is a transformer-based MIL framework that leverages both morphological and spatial correlations among instances to enhance visualization, interpretability, and performance in WSI pathology classification.
	\item \textbf{CLAM-SB} \cite{lu2021data} is a clustering constraint-based attention multiple instance learning method that employs a single attention branch to aggregate instance features and generate a bag-level representation.
	\item \textbf{CLAM-MB} \cite{lu2021data} is the multi-branch version of the CLAM model, computing attention scores for each class separately to produce multiple unique bag-level representations.
	\item \textbf{DTFD-MIL} \cite{zhang2022dtfd} addresses the challenge of limited WSI samples in MIL by introducing pseudo-bags to virtually enlarge the bag count and implementing a double-tier framework that leverages an attention-based derivation of instance probabilities to effectively utilize intrinsic features.
	\item \textbf{ACMIL} \cite{zhang2024attention} mitigates overfitting by employing multiple branch attention and stochastic top-K instance masking to reduce attention value concentration and capture more discriminative instances in WSI classification.
	\item \textbf{ILRA} \cite{xiang2023exploring} incorporates a pathology-specific Low-Rank Constraint for feature embedding and an iterative low-rank attention model for feature aggregation, achieving enhanced performance in gigapixel-sized WSI classification.
	\item \textbf{DGRMIL} \cite{zhu2024dgr}  models instance diversity by converting instance embeddings into similarities with predefined global vectors via a cross-attention mechanism and further enhances the diversity among these global vectors through positive instance alignment and a determinant point process-based diversified learning paradigm.
\end{itemize}

\begin{table*}[ht]
	\centering
	\resizebox{1\linewidth}{!}{\begin{tabular}{lccccccccccccccc}
		\toprule
		& \multicolumn{5}{c}{\textbf{STAS\_TCGA}} & \multicolumn{5}{c}{\textbf{STAS\_CPTAC}} & \multicolumn{5}{c}{\textbf{STAS\_CSU}} \\
		\cmidrule(lr){2-6}\cmidrule(lr){7-11}\cmidrule(lr){12-16}
		\textbf{Method} & \textbf{ACC} & \textbf{AUC} & \textbf{F1} & \textbf{Recall} & \textbf{Precision} 
		& \textbf{ACC} & \textbf{AUC} & \textbf{F1} & \textbf{Recall} & \textbf{Precision}
		& \textbf{ACC} & \textbf{AUC} & \textbf{F1} & \textbf{Recall} & \textbf{Precision} \\
		\midrule
		Maxpooling   & 0.5661 & 0.5201 & \textbf{0.5801} & 0.5904 & 0.5904 
		& 0.5850 & 0.5940 & 0.5667 & 0.5850   & 0.5850 
		& 0.5751 & \textbf{0.6055} & 0.5639 & 0.5751  & 0.5751 \\
		Meanpooling  & 0.5971 & 0.5845 & 0.5098 & 0.5971  & 0.5971 
		& 0.5875 & 0.6026 & 0.5502 & 0.5875  & 0.5875 
		& 0.5655 & 0.5963 & 0.5527 & 0.5655  & 0.5655 \\
		ABMIL        & 0.5991 & 0.5008 & 0.5151 & 0.5991  & 0.5991 
		& 0.5950 & 0.6379 & 0.5738 & 0.595   & 0.595 
		& 0.5582 & 0.5764 & 0.5582 & 0.5582  & 0.5582 \\
		TransMIL     & 0.5966 & \textbf{0.5926} & 0.5609 & 0.5966  & 0.5966 
		& 0.5925 & 0.5996 & 0.5707 & 0.5925  & 0.5925 
		& 0.5422 & 0.5569 & 0.5395 & 0.5422  & 0.5422 \\
		CLAM-SB     & 0.6017 & 0.5542 & 0.5262 & 0.6017  & 0.6017
		& 0.5800 & 0.5850 & 0.5508 & 0.6000  & 0.6000 
		& 0.5574 & 0.5730 & 0.5648 & 0.5574  & 0.5574 \\
		CLAM-MB      & 0.5658 & 0.5395 & 0.5044 & 0.5658  & 0.5658 
		& 0.6275 & 0.6136 & 0.6137 & 0.6275  & 0.6275 
		& 0.5687 & 0.5946 & 0.5633 & 0.5687  & 0.5687 \\
		DTFD-MIL         & 0.5900 & 0.4792 & 0.4434 & 0.5900  & 0.5900 
		& 0.6100 & 0.5033 & 0.5887 & 0.6100  & 0.6100 
		& \textbf{0.5767} & 0.5625 & 0.5742 & 0.5767  & 0.5767 \\
		ACMIL        & 0.5493 & 0.5754 & 0.5473 & 0.5493  & 0.5523
		& 0.5670 & 0.5889 & 0.5640 & 0.5670  & 0.5718
		& 0.5469 & 0.5851 & 0.5406 & 0.5469  & 0.5508 \\
		ILRA         & 0.4723 & 0.5271 & 0.4953 & 0.5414  & 0.6056
		& 0.5219 & 0.6239 & 0.6437 & \textbf{0.6939}  & \textbf{0.6765}
		& 0.5390 & 0.5967 & 0.6007 & 0.6368  & \textbf{0.6124} \\
		DGRMIL       & 0.5621 & 0.6084 & 0.5841 & 0.5963  & \textbf{0.6347}
		& 0.6050 & 0.6198 & 0.6230 & 0.6444  & 0.6401
		& 0.5116 & 0.5854 & \textbf{0.6048} & \textbf{0.651}  & 0.5939 \\
		SMILE(our)         & \textbf{0.6064} & 0.5736 & 0.5079 & \textbf{0.6064}  & 0.6064
		& \textbf{0.6450} & \textbf{0.6517} & \textbf{0.6242} & 0.645  & 0.6450
		& 0.5655 & 0.5979 & 0.5567 & 0.5655  & 0.5655 \\
		\bottomrule
	\end{tabular}}
	\caption{Baseline results of SOTA MIL and SMILE methods on STAS\_TCGA, STAS\_CPTAC and STAS\_CSU datasets.}
	\label{tab:comparison}
\end{table*}

\subsection{Results and Analysis}

Table~\ref{tab:comparison} compares SMILE with ten state-of-the-art (SOTA) models using evaluation metrics across the three proposed STAS datasets. All models learn weak labels of lung cancer WSI to diagnose STAS. Overall, we observed performance variations across the three datasets for all models.

	\noindent  \textbf{STAS\_TCGA:} 
	For accuracy, methods like CLAM-SB (0.6017) and ABMIL (0.5991) attain competitive results. Our method achieves an accuracy of 0.6064 and an AUC of 0.5736. Although the DGRMIL approach shows a slightly higher AUC (0.6084), our overall performance remains in a comparable range. This indicates that our approach can maintain reasonable accuracy while balancing the trade-off with AUC.
	
	\noindent  \textbf{STAS\_CPTAC:}
	SMILE achieves the highest accuracy (0.645) among all listed models and also yields a leading AUC value of 0.6517. The improvement in these two metrics highlights the advantages of the scale-adaptive attention mechanism and the MIL framework in learning STAS pathological features. By contrast, alternative methods (e.g., CLAM-MB, TransMIL, DGRMIL) show lower accuracy and AUC, highlighting that combining local instance features and global context can significantly enhance STAS classification.
	
	\noindent  \textbf{STAS\_CSU:}
	SMILE obtains an accuracy of 0.5655 with an AUC of 0.5979. Some methods (e.g., Maxpooling and DTFD-MIL) yield higher accuracy or AUC in certain cases, but not consistently across both metrics. Notably, CLAM-MB achieves accuracy = 0.5687 (higher than ours by a small margin), though with a similar AUC (0.5946). Overall, the results suggest that further refinements or domain adaptation strategies might improve the robustness of our model on this particular dataset.

In summary, the proposed SMILE achieves the highest combined accuracy and AUC performance on STAS\_CPTAC, and competitive results on STAS\_TCGA and STAS\_CSU. Given the complexities of STAS classification and the variability of histopathological data, these results demonstrate that incorporating multi-instance learning with a scale-adaptive attention mechanism can effectively capture the fine-grained patterns essential for STAS diagnosis. Future work may focus on improving domain adaptation and interpretability to further enhance the model’s performance across diverse datasets.

\subsection{Ablation Study}

In this section, we explore the impact of different thresholds and scaling factors on model performance. We conducted experiments on three representative datasets and recorded changes in key metrics including accuracy, AUC, and F1 scores under various threshold and scale-factor settings. To ensure the reliability of our results, cross-validation was employed on each dataset to reduce potential biases stemming from data splits. Table~\ref{tab:ablation} presents the experimental findings.

\begin{table}[ht]
	\centering
	\resizebox{1\linewidth}{!}{\begin{tabular}{cccccccc}
		\toprule
		\textbf{Threshold} & \textbf{Factor} & \multicolumn{3}{c}{\textbf{STAS\_TCGA}} & \multicolumn{3}{c}{\textbf{STAS\_CPTAC}} \\
		\cmidrule(lr){3-5}\cmidrule(lr){6-8}
		&  & \textbf{Acc} & \textbf{AUC} & \textbf{F1} & \textbf{Acc} & \textbf{AUC} & \textbf{F1} \\
		\midrule
		\textbf{w/o} & \textbf{w/o} & 0.6091 & 0.5595 & 0.5190 & 0.6325 & 0.6595 & 0.6129 \\
		\cmidrule(lr){1-8}
		0.5 & 0.3 & 0.6137 & \textbf{0.5841} & 0.5148 & 0.6425 & 0.6531 & 0.6240 \\
		& 0.4 & 0.6064 & 0.5747 & 0.5056 & 0.6350 & 0.6560 & 0.6155 \\
		& 0.5 & 0.6064 & 0.5736 & 0.5079 & \textbf{0.6450} & 0.6517 & \textbf{0.6242} \\
		& 0.6 & \textbf{0.6164} & 0.5536 & \textbf{0.5368} & 0.6325 & 0.6574 & 0.6123 \\
		& 0.7 & 0.6041 & 0.5693 & 0.5227 & 0.6375 & \textbf{0.6586} & 0.6149 \\
		& 0.8 & 0.6042 & 0.5631 & 0.5156 & 0.6250 & 0.6551 & 0.6036 \\
		\cmidrule(lr){1-8}
		0.6 & 0.3 & \textbf{0.6112} & \textbf{0.5703} & \textbf{0.5295} & 0.6350 & 0.6523 & 0.6150 \\
		& 0.4 & 0.5898 & 0.5627 & 0.4838 & 0.6350 & 0.6467 & 0.6127 \\
		& 0.5 & 0.6039 & 0.5611 & 0.5106 & \textbf{0.6450} & 0.6526 & \textbf{0.6245} \\
		& 0.6 & 0.5967 & 0.5619 & 0.5102 & 0.6400 & 0.6569 & 0.6196 \\
		& 0.7 & 0.6016 & 0.5650 & 0.5114 & 0.6300 & \textbf{0.6600} & 0.6091 \\
		& 0.8 & 0.5971 & 0.5603 & 0.5029 & 0.6350 & 0.6552 & 0.6141 \\
		\cmidrule(lr){1-8}
		0.8 & 0.3 & 0.5969 & \textbf{0.5757} & 0.5106 & 0.6325 & 0.6414 & 0.6115 \\
		& 0.4 & 0.6064 & 0.5691 & \textbf{0.5179} & \textbf{0.6425} & 0.6473 & \textbf{0.6227} \\
		& 0.5 & \textbf{0.6089} & 0.5667 & \textbf{0.5179} & 0.6400 & 0.6520 & 0.6204 \\
		& 0.6 & 0.6041 & 0.5624 & 0.5167 & 0.6375 & 0.6538 & 0.6182 \\
		& 0.7 & 0.5969 & 0.5654 & 0.5028 & 0.6400 & 0.6569 & 0.6200 \\
		& 0.8 & 0.6042 & 0.5589 & 0.5031 & 0.6325 & \textbf{0.6587} & 0.6132 \\
		\bottomrule
	\end{tabular}}
		\caption{The impact of different scale factors and thresholds on model performance (w/o indicates no threshold or scale-factor adjustments).}
		\label{tab:ablation}
\end{table}

By examining the data in Table~\ref{tab:ablation}, we find that thresholds and scaling factors significantly and intricately affect the model's performance. In particular, for the STAS\_TCGA and STAS\_CPTAC datasets, accuracy, AUC, and F1 tend to increase when we reduce threshold and scaling-factor values. When the threshold is 0.5 and the scale factor is 0.5, our method reaches a local optimum on STAS\_CPTAC, whereas for STAS\_TCGA, a local optimum is observed at threshold = 0.5 and factor = 0.6. These results indicate that striking a balance between threshold-based filtering and scaling-based adjustments can effectively boost model performance.

\section{Discussion and Conclusion}
In this study, we propose a \emph{scale-aware} multiple instance learning  framework to address the challenges of diagnosing STAS in lung cancer WSIs. Our experiments conducted on the STAS\_TCGA, STAS\_CPTAC, and STAS\_CSU datasets demonstrate that scale-adaptive attention-based feature aggregation significantly improves diagnostic performance. The proposed method achieves notable accuracy (0.6450) and competitive AUC (0.6517) on the STAS\_CPTAC, highlighting the effectiveness of our approach in capturing STAS patterns within heterogeneous histopathological images. Ablation studies reveal that both threshold and scaling factors substantially impact performance metrics. This finding emphasizes the importance of tailoring hyperparameters for different datasets, considering inherent variability in image quality, staining conditions, and pathological subtypes. Furthermore, scale-adaptive adjustment mitigates the issue of over-attention to high-salience instances. By selectively scaling attention scores above thresholds, our method achieves more balanced focus on subtle and prominent local features, enhancing its robustness in identifying sparse or heterogeneous STAS manifestations. 

Despite these positive outcomes, two main limitations persist. First, the proposed model shows slightly inferior performance on the STAS\_CSU compared to STAS\_CPTAC, suggesting the need for domain adaptation or data augmentation. Second, while the scale-adaptive mechanism improves diagnostic accuracy compared to SOTA MIL approaches, further research is required to bridge the gap between visual heatmaps and precise pathologist-level annotations. The complex pathological features of STAS underscore the importance of more refined and interpretable modeling techniques, potentially involving graph-based WSI representations.




\bibliographystyle{named}
\bibliography{ijcai25}
\end{CJK}
\end{document}